\documentclass[conference]{IEEEtran}
\IEEEoverridecommandlockouts
\usepackage{cite}
\usepackage{amsmath,amssymb,amsfonts}
\usepackage{algorithmic}
\usepackage{graphicx}
\usepackage{textcomp}
\usepackage{xcolor}
\usepackage{xcolor}
\usepackage[colorlinks=true,
            linkcolor=blue,
            citecolor=blue,
            urlcolor=blue]{hyperref}
\usepackage{doi}
\usepackage{url}
\usepackage{cleveref}
\crefname{figure}{Fig.}{Figs.}
\Crefname{figure}{Fig.}{Figs.}

\def\BibTeX{{\rm B\kern-.05em{\sc i\kern-.025em b}\kern-.08em
    T\kern-.1667em\lower.7ex\hbox{E}\kern-.125emX}}
\begin{document}

\title{Bridging the Gap: Toward Cognitive Autonomy
in Artificial Intelligence\\
\thanks{Identify applicable funding agency here. If none, delete this.}
}

\title{Bridging the Gap: Toward Cognitive Autonomy
in Artificial Intelligence}

\author{
\IEEEauthorblockN{
Noorbakhsh Amiri Golilarz,
Sindhuja Penchala,
Shahram Rahimi
}
\IEEEauthorblockA{
Department of Computer Science, The University of Alabama, Tuscaloosa, AL, USA\\
\{noor.amiri, srahimi1\}@ua.edu,\; spenchala@crimson.ua.edu
}
}


\maketitle

\begin{abstract}
Artificial intelligence has advanced rapidly across perception, language, reasoning, and multimodal domains. Yet despite these achievements, modern AI systems remain fundamentally limited in their ability to self-monitor, self-correct, and regulate their behavior autonomously in dynamic contexts. This paper identifies and analyzes seven core deficiencies that constrain contemporary AI models: the absence of intrinsic self-monitoring, lack of meta-cognitive awareness, fixed and non-adaptive learning mechanisms, inability to restructure goals, lack of representational maintenance, insufficient embodied feedback, and the absence of intrinsic agency. Alongside identifying these limitations, we also outline a forward-looking perspective on how AI may evolve beyond them through architectures that mirror neurocognitive principles. We argue that these structural limitations prevent current architectures, including deep learning and transformer-based systems, from achieving robust generalization, lifelong adaptability, and real-world autonomy. Drawing on a comparative analysis of artificial systems and biological cognition \cite{golilarz2025towards}, and integrating insights from AI research, cognitive science, and neuroscience, we outline how these capabilities are absent in current models and why scaling alone cannot resolve them. We conclude by advocating for a paradigmatic shift toward cognitively grounded AI (cognitive autonomy) capable of self-directed adaptation, dynamic representation management, and intentional, goal-oriented behavior, paired with reformative oversight mechanisms \cite{golilarz2025reforming} that ensure autonomous systems remain interpretable, governable, and aligned with human values.
\end{abstract}

\begin{IEEEkeywords}
Artificial intelligence, self-monitoring, biological cognition, cognitive autonomy.
\end{IEEEkeywords}

\section{Introduction}
\label{sec: intro}

Artificial intelligence has undergone a period of rapid acceleration driven chiefly by advances in deep learning and transformer architectures, enabling breakthroughs across language modeling, computer vision, scientific discovery, and multimodal reasoning \cite{vaswani2017attention} \cite{radford2021learning} \cite{mann2020language}. Despite this progress, contemporary AI systems continue to exhibit fundamental limitations in autonomy, adaptability, and self-regulation. While large models can generate fluent language, solve complex recognition tasks, and perform sophisticated reasoning under some constraints, their capabilities remain structurally bounded by a static learning paradigm, the absence of intrinsic self-evaluation, and their dependency on externally imposed objectives \cite{marcus2002next} \cite{lake2017building}. 

In contrast, biologically-grounded intelligence demonstrates continual self-assessment, context-dependent strategy adjustment, adaptive learning across multiple timescales, and embodied, exploratory capability that unfolds without external supervision \cite{lin2021truthfulqa} \cite{fabiano2025thinking}  \cite{johnsonimagining}. Metacognitive monitoring in humans, supported by prefrontal circuitry, enables the brain to evaluate confidence in its own decisions and adjust behavior accordingly \cite{fleming2012neural}. Learning itself is distributed across fast and slow timescales, with short-term and long-term synaptic plasticity jointly supporting rapid adaptation and stable knowledge acquisition \cite{o2023nonlinear}. At the level of perception and action, predictive processing accounts of the brain emphasize closed perception–action loops \cite{golilarz2025towards} in which agents actively sample and reshape their environment to minimize prediction error, rather than passively receiving inputs \cite{clark2015embodied}. Moreover, intrinsic motivation and curiosity drive spontaneous exploration and open-ended skill acquisition, providing an internal engine for self-directed learning even in the absence of explicit external rewards \cite{oudeyer2007intrinsic}.

Artificial intelligence systems, by comparison, operate in a narrow reactive mode. This contrast raises a foundational question: what essential components of cognition are missing from today’s AI systems to achieve cognitive autonomy? The present paper addresses this question directly. We argue that there exist foundational deficiencies in contemporary AI architectures that prevent current systems from achieving robust autonomy, human-like adaptability, and cognitively grounded behavior. These deficiencies relate to missing capacities in self-monitoring, meta-awareness, adaptive plasticity, goal restructuring, representational repair, embodied feedback, and autonomous initiative. We discuss these deficiencies conceptually, grounding each in both AI practice and cognitive science theory, and we organize them into coherent structural dimensions that reveal why modern AI succeeds spectacularly in narrow contexts yet fails in open, dynamic, and uncertain environments. Our objective is not merely to critique the current paradigm, but to articulate a conceptual foundation for the next stage of AI research, one in which systems must not simply perform, but understand, regulate, and evolve themselves.

The rest of the paper is organized as follows. \Cref{sec:limit} provides a detailed examination of the seven foundational cognitive capacities missing in contemporary AI systems. \Cref{sec:implication} analyzes the broader implications of these deficiencies for reliability, generalization, lifelong adaptation, and the emergence of autonomous cognitive behavior. \Cref{sec:future} outlines future research directions, highlighting how neurocognitive-inspired architectures may offer a pathway toward overcoming these structural limitations and enabling cognitive autonomy. \Cref{sec:conclusion} concludes with a synthesis of key insights and discusses how closing these gaps may redefine the trajectory of artificial intelligence.

\section{What Is Missing in Today’s AI?}
\label{sec:limit}

\begin{figure}[htbp]
    \centering
    \includegraphics[width=1.0\linewidth]{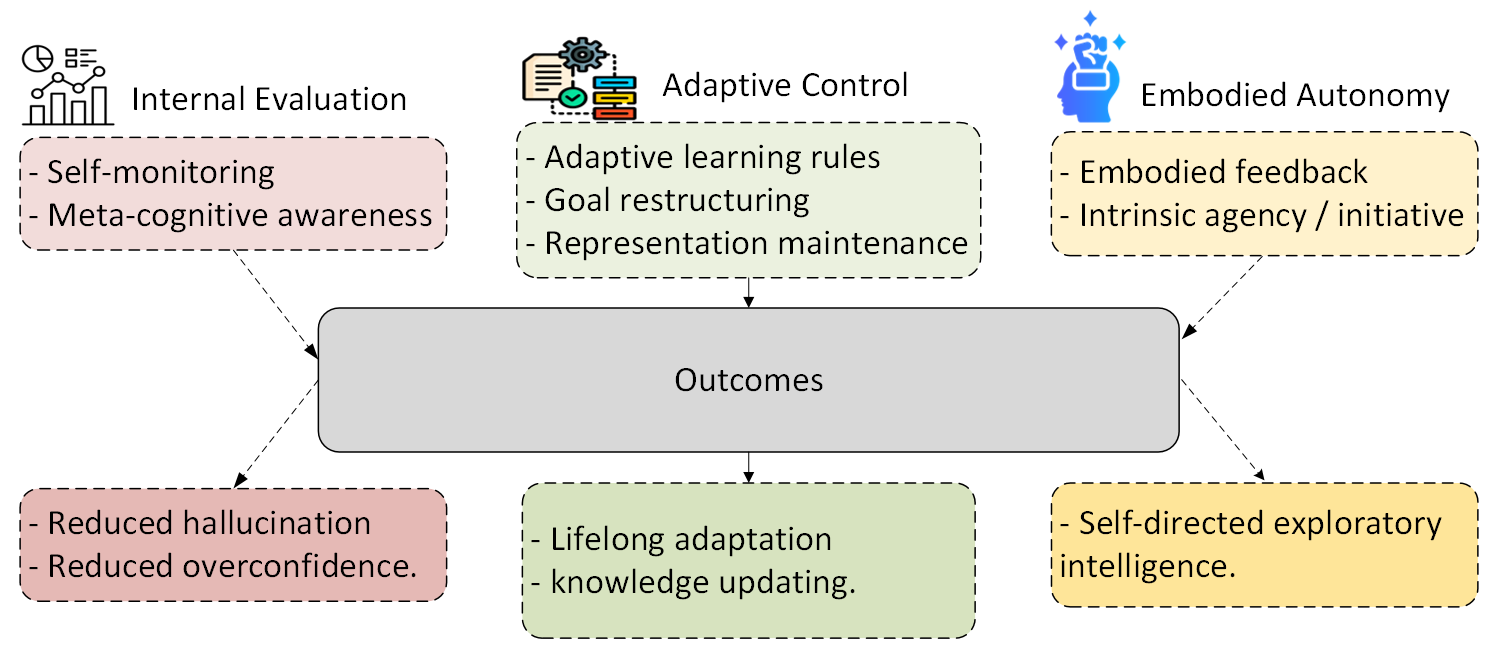}
    \caption{Structural organization of seven foundational cognitive capacities missing in current AI architectures, categorized into three functional domains: internal evaluation, adaptive control, and embodied autonomy. Together, these capabilities enable the emergence of cognitive autonomy, which in turn manifests as more reliable, adaptive, and self-directed intelligence.}
    \label{fig:Structural}
\end{figure}

The limitations outlined in this section reflect structural rather than incremental shortcomings of contemporary machine intelligence. Despite unprecedented technical progress, modern AI systems exhibit deep deficiencies in self-evaluation, adaptive learning, representational continuity, and autonomous cognitive control. These limitations are not isolated, but interdependent and systematically rooted in the architecture of current AI systems, forming a coherent pattern of missing cognitive capacities. To clarify this structure, we organize these deficiencies into three functional domains and illustrate their relationships conceptually in \Cref{fig:Structural}, before examining each capacity in detail.

Unlike biological cognition, which integrates perception, memory, adaptation, and agency into a unified continuous loop, current models remain constrained by static training regimes, externally imposed objectives, and rigid inference procedures. These deficiencies undermine not only performance robustness, but also the potential for safe deployment, flexible generalization, and autonomous operation. To contextualize these shortcomings,  \Cref{tab:cognitive_capacities} summarizes the foundational capacities absent in modern AI and their corresponding cognitive roles in biological systems, serving as a structural map for the subsections that follow.

\begin{table}[htbp]
\centering
\caption{Foundational Cognitive Capacities Missing in Modern AI and Their Biological Counterparts}
\vspace{-0.2cm}
\label{tab:cognitive_capacities}
\setlength{\tabcolsep}{3pt}        
\renewcommand{\arraystretch}{1.1}  

\begin{tabular}{p{0.26\columnwidth} p{0.38\columnwidth} p{0.30\columnwidth}}
\hline
\textbf{Missing Capacity in AI} &
\textbf{Associated Cognitive Function in Biological Systems} &
\textbf{Primary Consequence in AI} \\
\hline
Self-monitoring        & Confidence evaluation, error awareness           & Undetected failures and hallucinations \\
Meta-cognitive awareness & Knowledge boundary tracking                     & Inability to assess knowledge gaps \\
Adaptive learning rules & Contextual and plasticity-based learning                  & No online adaptation or rapid learning \\
Goal restructuring     & Dynamic priority adjustment                      & Rigid, externally fixed goal execution \\
Representation repair  & Memory reconsolidation and restructuring         & Accumulating representational damage \\
Embodied feedback      & Sensorimotor coupling                            & No refinement through experience \\
Agency / initiative    & Intrinsic motivation \& exploration              & Passive, reactive problem solving \\
\hline
\end{tabular}
\end{table}



\subsection{Self-Monitoring}
Contemporary AI models lack the capacity for intrinsic self-monitoring, the ability to detect internal inconsistency, uncertainty, or error during inference without external oversight. Unlike biological cognition, where metacognitive monitoring circuits within prefrontal cortical systems continuously evaluate confidence and coherence in decision-making \cite{fleming2012neural}, modern neural networks possess no endogenous self-evaluative mechanisms. They do not internally signal epistemic uncertainty, nor can they distinguish between reliable and unreliable outputs. As a result, when transformers and large language models produce hallucinated responses, they do so with unjustified confidence because they have no structural process for identifying representational conflict or logical implausibility in their reasoning. This deficiency yields brittle behavior under slight distributional shifts, system overconfidence, and failure modes that are systematically invisible to the system itself because the detection of error remains an external function imposed during offline evaluation rather than a built-in, continuously engaged cognitive function.

\subsection{Meta-Cognitive Awareness}

While humans maintain meta-cognitive awareness of their own knowledge state, reflectively tracking what they know, what they do not know, and what information they must acquire next, modern AI systems lack the ability to self-interrogate the boundaries of their own competence. Transformers operate according to fixed computational pathways that do not evaluate the adequacy of internal representations, identify conceptual gaps, or prioritize learning based on uncertainty, novelty, or prediction error \cite{schaeffer2021algorithmic}. This absence of meta-awareness prevents AI models from strategically allocating computational resources, seeking clarifying information, or modulating their own learning and inference behavior in accordance with their confidence. In practice, AI systems do not recognize when additional information is necessary and do not distinguish between domains where they are competent and domains where they are structurally unprepared. Without the ability to model their own ignorance, these systems cannot meaningfully advance toward autonomous, self-directed learning.

\subsection{Adaptive Learning Rules}

A central limitation of contemporary AI systems is that their learning rules are entirely static once training ends. Deep networks are optimized with fixed gradient-based procedures during an offline training phase, and after deployment, their parameters and their plasticity cease to evolve. This stands in stark contrast to biological learning, which incorporates rapid synaptic plasticity, multi-timescale memory formation, and neuromodulatory learning mechanisms that support continuous adaptation, contextual learning, and episodic acquisition \cite{abbott2000synaptic} \cite{oja1982simplified}. Because AI learning mechanisms are rigid and externally defined, models cannot modify the way they learn in response to new goals, failures, uncertainty, or environmental change. There is no mechanism for updating plasticity rules, adopting new adaptation strategies, or converting short-term interaction into structural learning. This results in systems that remain frozen with respect to learning dynamics, fundamentally unable to alter how they learn or restructure learning processes autonomously.

\subsection{Goal Restructuring}

Current AI systems cannot generate, adjust, or abandon goals based on internal evaluation or changing environmental conditions. Goals exist only as externally defined prompts, reward functions, or task specifications. In contrast, biological cognition constantly reshapes goal hierarchies and motivational priorities based on risk, reward probability, failure accumulation, resource constraints, and shifting contextual demands \cite{locke2010motivational}. AI systems lack mechanisms for modifying goal orientation at runtime, resolving conflicting objectives, or reorganizing behavioral strategies in accordance with internal or external feedback. This makes existing architectures inherently reactive rather than strategic. Instead of planning adaptively as circumstances evolve, they simply execute pre-defined target conditions. The absence of goal restructuring precludes autonomous decision-making and limits the system’s capacity to behave in a self-directed, intentionally flexible manner.

\subsection{Representation Repair}
Deep neural networks are fundamentally incapable of repairing or restructuring their internal representations when they degrade, conflict, or become outdated. Representation drift, inconsistency, or fragmentation accumulates over time without internal correction. Biological cognition, in contrast, maintains stability and coherence through mechanisms such as reconsolidation, schema reorganization, and continuous memory restructuring \cite{dudai2012restless}. Modern AI systems have no such mechanisms: once a latent space becomes incoherent, no endogenous corrective process realigns the learned representation. This lack of representational maintenance leads to compounding errors, semantic fragmentation, and degraded generalization capacity. In effect, models accumulate representational damage that they cannot detect and cannot repair.

\subsection{Embodied Feedback}
Even as models grow increasingly capable of symbolic reasoning and pattern recognition, the overwhelming majority of contemporary AI systems still lack meaningful interaction with the physical or sensory world. As a result, they do not learn through direct environmental feedback, do not experience the consequence of their errors, and do not refine behavior through embodied exploration. Human learning emerges from tightly coupled perception–action loops, where success, error, and uncertainty are continuously evaluated through sensory-motor engagement \cite{clark2015surfing}. Models trained solely on static datasets or synthetic corpora lack sensorimotor grounding and therefore cannot acquire the type of fine-grained adaptive competence necessary for real-world intelligence. Without embodied feedback, AI systems cannot meaningfully test hypotheses, validate assumptions, or refine internal models based on lived interaction.

\subsection{Agency or Initiative}
Finally, contemporary AI lacks agency, the capacity to autonomously initiate action, exploration, inquiry, or hypothesis formation. AI systems do not act unless instructed; they possess no intrinsic motivation to reduce uncertainty, improve understanding, or collect missing information. In contrast, biological intelligence is driven by internally generated motivation systems linked to curiosity, exploratory drive, and intrinsic reward mechanisms \cite{northoff2025brain} \cite{papesh2012memory}. The absence of agency confines AI systems to passive execution modes, preventing them from engaging proactively with tasks, environments, or knowledge gaps. Without agency, AI cannot develop into a system capable of true autonomy, self-guided learning, or self-initiated progress.

\section{Implications For Future AI Systems Development}
\label{sec:implication}

The deficiencies identified in \Cref{sec:limit} reveal deep conceptual gaps that restrict modern AI systems from achieving robust, autonomous performance in real-world environments. These gaps manifest not only as technical limitations but as structural barriers that undermine reliability, adaptability, safety, and cognitive scalability. \Cref{Tab:implications} outlines how the seven deficiencies translate into three high-level implication domains: reliability and safety, generalization and lifelong learning, and autonomous cognitive behavior. Each implication reflects a dimension in which current AI systems fail to match even basic characteristics of biological cognition, thereby constraining their capacity to operate flexibly and independently in open-ended contexts.

\begin{table}[htbp]
\centering
\caption{High-Level Implications of Missing Cognitive Capacities in Modern AI}
\vspace{-0.2cm}
\label{Tab:implications}

\setlength{\tabcolsep}{3pt}        
\renewcommand{\arraystretch}{1.1}  

\begin{tabular}{p{0.25\columnwidth} p{0.35\columnwidth} p{0.35\columnwidth}}
\hline
\textbf{Implication Domain} &
\textbf{Underlying Deficiencies} &
\textbf{Resulting Limitations} \\
\hline
Reliability \& Safety &
Lack of self-monitoring; lack of representation repair &
Overconfident errors, undetected failures, hallucinations, safety-critical brittleness \\
Generalization \& Lifelong Adaptation &
No adaptive learning rules; no embodied feedback; no goal restructuring &
Poor transfer learning, inability to adapt online, rigid behaviors, limited generalization beyond training distributions \\
Autonomous Cognitive Behavior &
No meta-cognitive awareness; no intrinsic agency/initiative &
No self-directed exploration, no strategic planning, inability to form or revise goals, absence of autonomous reasoning \\
\hline
\end{tabular}
\end{table}

\subsection{Reliability and Safety Implications}
The lack of intrinsic self-monitoring and representational repair mechanisms severely undermines the reliability of current AI systems. Models that cannot detect internal inconsistency or uncertainty become vulnerable to overconfident failures and hallucinations, especially in high-stakes environments such as healthcare, autonomous driving, and industrial safety. Without internal error recognition, AI systems mask their failures rather than attenuate them, creating a fundamental safety risk in contexts where external oversight cannot be guaranteed.

\subsection{Limitations in Generalization and Lifelong Adaptation}
Modern AI systems exhibit limited capacity to generalize beyond their training distributions because they lack adaptive learning rules, embodied feedback loops, and the ability to restructure goals. Unlike biological agents that continually reshape strategies in response to context, deep models remain bound to static policies defined during pretraining. This rigidity inhibits transfer, slows adaptation to new conditions, and prevents the incremental accumulation of knowledge over a system’s lifetime. As summarized in \Cref{Tab:implications}, these limitations directly restrict the scalability of AI in dynamic environments.

\subsection{Barriers to Autonomous Cognitive Behavior}
Perhaps most significantly, the absence of meta-cognitive awareness and intrinsic agency prevents the emergence of genuinely autonomous cognitive behavior. AI systems today cannot initiate exploration, form new goals, or revise existing objectives without explicit external instruction, restricting them to reactive rather than proactive operation. This absence of agency blocks progress toward systems capable of independent hypothesis formation, curiosity-driven learning, or self-directed knowledge expansion, all of which are central to autonomous intelligence.

\begin{figure}[htbp]
    \centering
    \includegraphics[width=1.0\linewidth]{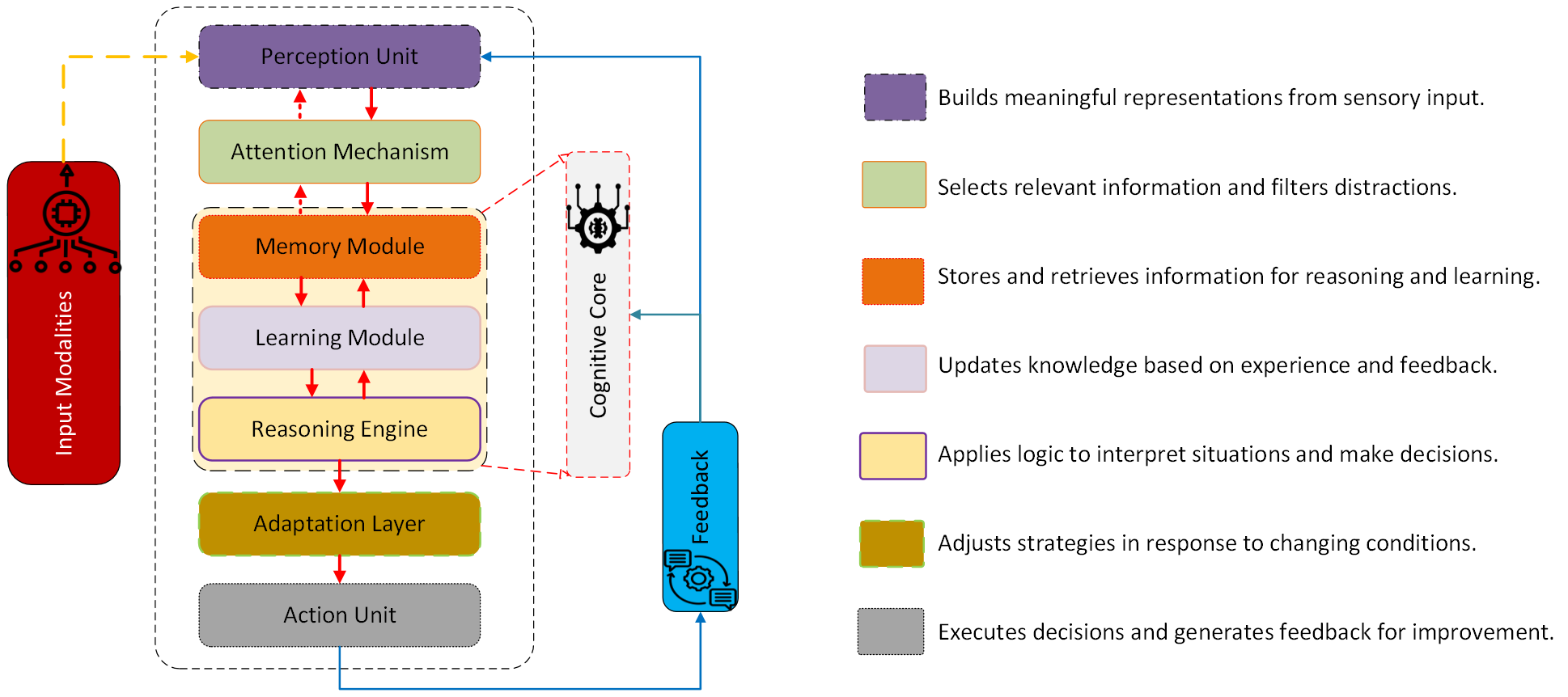}
    \caption{Conceptual closed-loop cognitive architecture illustrating the core components for autonomous, neurocognitive-inspired intelligence \cite{golilarz2025towards}. The system integrates perception, attention, memory, learning, reasoning, and adaptive control into a unified cognitive cycle that transforms input into action, and uses environmental feedback to continuously refine internal representations, strategies, and decisions.}
    \label{fig:Conceptual}
\end{figure}

\section{Future Directions: Toward Cognitive Autonomy}
\label{sec:future}

The analysis presented in this paper identifies structural cognitive capacities that are absent from contemporary AI systems. Bridging these gaps will require architectural and algorithmic innovations that move beyond pattern recognition toward systems capable of self-assessment, adaptive learning, representational maintenance, and autonomous goal management. As a forward-looking direction, recent work in neurocognitive-inspired intelligence \cite{golilarz2025towards} suggests the potential of integrating these capacities into a unified cognitive architecture (\Cref{fig:Conceptual}). Rather than operating as static function approximators, such systems would be organized around a closed-loop cognitive cycle that includes perception, attention, memory, learning, reasoning, adaptation, and action. The aim of this conceptual blueprint is not to prescribe a specific engineered solution, but to illustrate the type of system-level organization required for cognitive autonomy. A unified cognitive loop could support continuous uncertainty monitoring, dynamic adjustment of learning rules, goal restructuring, and interaction-driven refinement, characteristics that define biological intelligence but remain inaccessible to current machine learning models. By aligning future AI systems with these principles, researchers may begin to close the gap between engineered performance and autonomous cognition, opening pathways toward more adaptive, resilient, and self-directed artificial agents.

A promising avenue for progress lies in the development of adaptive learning mechanisms that can modify themselves at runtime. Current deep models treat learning as a phase that ends at deployment; future systems must treat learning as a persistent and evolving process. This suggests the need for plasticity-aware optimization frameworks capable of updating not only weights, but learning rules themselves. Approaches such as meta-learning, neuromodulatory control, and memory-consolidation-inspired mechanisms could support incremental, context-dependent knowledge acquisition. A crucial research challenge is enabling models to distinguish between information that should be transient and information that should be structurally encoded, a capability analogous to synaptic tagging, consolidation, and reconsolidation in biological brains. If achieved, this would allow artificial agents to refine their understanding continuously, repair their own internal representations, and adapt dynamically without catastrophic forgetting or retraining from scratch.

Another critical direction involves embedding intrinsic motivation and goal-forming capabilities. Present-day systems act only when prompted, but autonomous intelligence requires self-initiated exploration and the capacity to set internal objectives. This raises several technical and philosophical challenges: How should an AI system generate new goals? By what criteria should it abandon them? What mechanisms ensure that internally generated objectives remain aligned with safety, coherence, and long-term utility? Inspiration may be drawn from models of curiosity-driven learning, reward-free exploration, and hierarchical control in biological organisms. Future research should investigate architectures in which uncertainty, novelty detection, and prediction error serve as internal drive signals, enabling agents not only to respond to tasks, but to seek them. Such systems would no longer be passive pattern processors, but active cognitive entities capable of forming hypotheses, testing them in the world, and revising internal goals through experience.

Ultimately, achieving cognitive autonomy will require a synthesis of developments across multiple fronts: adaptive plasticity and self-reflective learning, representational repair, embodied interaction, and agentic goal-formation. The path forward is challenging, but conceptually well-mapped. If these capacities can be integrated into a coherent architecture, artificial intelligence may progress beyond static inference toward continuous self-directed adaptation, closing the conceptual gap between engineered systems and biological minds. The trajectory outlined here does not constitute an incremental extension of current methods, but a shift in how intelligence is constructed: from systems trained to perform tasks, to systems designed to understand, evaluate, and evolve themselves.

Achieving cognitive autonomy may raise a critical question: if we succeed, what kind of intelligence are we creating, and how do we remain its guide? As models gain the ability to self-monitor, revise goals, adapt learning rules, and repair internal representations, autonomy is no longer a technical milestone, it becomes a governance frontier. At the highest level, building minds also requires building boundaries. In our related work on reformative and cognitively contained AI \cite{golilarz2025reforming}, we argue that progress toward autonomous intelligence must be matched by equal progress in self-regulation, oversight, and cognitive firewalls that constrain reasoning trajectories when necessary. The goal is not to limit intelligence, but to ensure that it grows with insight rather than momentum, balancing agency with accountability, creativity with caution, and power with reflection.

\section{Conclusion}
\label{sec:conclusion}

Although contemporary AI systems have achieved extraordinary technical success, they remain fundamentally incomplete as cognitive entities. The seven deficiencies described in this analysis highlight structural limitations that preclude self-regulated, adaptive, and autonomous intelligence. At present, AI systems perceive without grounding, reason largely without reflection, and act without intention. They are powerful engines of pattern computation, yet fragile systems of cognition. Bridging this divide requires rethinking AI not as a static predictor but as a dynamic, self-maintaining, self-modifying organism embedded within a world it continuously interprets. The path forward, therefore, demands more than scaling models or expanding datasets. It requires integrating cognitive principles that enable systems to evaluate their own uncertainty, regulate learning, restructure internal representations, and initiate action in pursuit of meaningful goals. Achieving this transition may mark the emergence of a new phase in AI, one in which intelligence is not merely engineered, but cultivated. Looking ahead, progress toward cognitive autonomy will depend on unifying these principles into cohesive frameworks capable of continual self-assessment and context-sensitive adaptation. This transition represents not merely a technical challenge, but a conceptual transformation in how we understand and construct intelligent systems. Achieving it may mark the emergence of a new phase in artificial intelligence, one in which intelligence is not only engineered, but systematically cultivated through cognitive design.

\vspace{0.2cm}

\noindent\textbf{Acknowledgement:}
This work was supported by the BRAINS Lab (Bioinspired Robotics, AI, Imaging \& Neurocognitive Systems Laboratory) in the Department of Computer Science at The University of Alabama.


\begin{thebibliography}{00}

\bibitem{abbott2000synaptic}
L.~F. Abbott and S.~B. Nelson, ``Synaptic plasticity: Taming the beast,'' \emph{Nature Neuroscience}, vol.~3, no.~11, pp. 1178--1183, 2000. doi: 10.1038/81453.

\bibitem{clark2015embodied}
A.~Clark, ``Embodied prediction,'' in \emph{Open Mind}. Frankfurt am Main: MIND Group, 2015.

\bibitem{clark2015surfing}
A.~Clark, \emph{Surfing Uncertainty: Prediction, Action, and the Embodied Mind}. Oxford, U.K.: Oxford Univ. Press, 2015. doi: 10.1093/acprof:oso/9780190217013.001.0001.

\bibitem{dudai2012restless}
Y.~Dudai, ``The restless engram: Consolidations never end,'' \emph{Annual Review of Neuroscience}, vol.~35, pp. 227--247, 2012. doi: 10.1146/annurev-neuro-062111-150500.

\bibitem{fabiano2025thinking}
F.~Fabiano \emph{et al.}, ``Thinking fast and slow in human and machine intelligence,'' \emph{Communications of the ACM}, vol.~68, no.~8, pp. 72--79, 2025. doi: 10.1145/3715709.

\bibitem{fleming2012neural}
S.~M. Fleming and R.~J. Dolan, ``The neural basis of metacognitive ability,'' \emph{Philosophical Transactions of the Royal Society B}, vol.~367, no.~1594, pp. 1338--1349, 2012. doi: 10.1098/rstb.2011.0417.

\bibitem{golilarz2025towards}
N.~A. Golilarz, H.~S.~A. Khatib, and S.~Rahimi, ``Towards neurocognitive-inspired intelligence: From AI's structural mimicry to human-like functional cognition,'' arXiv:2510.13826, 2025.

\bibitem{golilarz2025reforming}
N.~A. Golilarz, H.~S.~A. Khatib, and S.~Rahimi, ``Reforming artificial intelligence: A call for cognitive containment,'' \emph{Preprints}, 2025. doi: 10.20944/preprints202511.0867.v1.

\bibitem{johnsonimagining}
S.~G.~B. Johnson \emph{et al.}, ``Imagining and building wise machines: The centrality of AI metacognition,'' arXiv:2411.02478, 2025.

\bibitem{lake2017building}
B.~M. Lake, T.~D. Ullman, J.~B. Tenenbaum, and S.~J. Gershman, ``Building machines that learn and think like people,'' \emph{Behavioral and Brain Sciences}, vol.~40, p. e253, 2017. doi: 10.1017/S0140525X16001837.

\bibitem{lin2021truthfulqa}
S.~Lin, J.~Hilton, and O.~Evans, ``TruthfulQA: Measuring how models mimic human falsehoods,'' arXiv:2109.07958, 2021.

\bibitem{locke2010motivational}
H.~S. Locke and T.~S. Braver, ``Motivational influences on cognitive control: A cognitive neuroscience perspective,'' in \emph{Self Control in Society, Mind, and Brain}. Oxford, U.K.: Oxford Univ. Press, 2010. doi: 10.1093/acprof:oso/9780195391381.003.0007.

\bibitem{mann2020language}
B.~Mann \emph{et al.}, ``Language models are few-shot learners,'' arXiv:2005.14165, 2020.

\bibitem{marcus2002next}
G.~Marcus, ``The next decade in AI: Four steps towards robust artificial intelligence,'' arXiv:2002.06177, 2020.

\bibitem{northoff2025brain}
G.~Northoff, A.~Wolman, and J.~Zhang, ``Brain dynamics shape cognition—Spatiotemporal neuroscience,'' \emph{Physics of Life Reviews}, 2025. doi: 10.1016/j.plrev.2025.07.009.

\bibitem{o2023nonlinear}
C.~O'Donnell, ``Nonlinear slow-timescale mechanisms in synaptic plasticity,'' \emph{Current Opinion in Neurobiology}, vol.~82, p. 102778, 2023. doi: 10.1016/j.conb.2023.102778.

\bibitem{oja1982simplified}
E.~Oja, ``Simplified neuron model as a principal component analyzer,'' \emph{Journal of Mathematical Biology}, vol.~15, no.~3, pp. 267--273, 1982. doi: 10.1007/BF00275687.

\bibitem{oudeyer2007intrinsic}
P.-Y. Oudeyer and F.~Kaplan, ``What is intrinsic motivation? A typology of computational approaches,'' \emph{Frontiers in Neurorobotics}, vol.~1, p. 108, 2007. doi: 10.3389/neuro.12.006.2007.

\bibitem{papesh2012memory}
M.~H. Papesh and S.~D. Goldinger, ``Memory in motion: Movement dynamics reveal memory strength,'' \emph{Psychonomic Bulletin \& Review}, vol.~19, no.~5, pp. 906--913, 2012. doi: 10.3758/s13423-012-0281-3.

\bibitem{radford2021learning}
A.~Radford \emph{et al.}, ``Learning transferable visual models from natural language supervision,'' in \emph{Proceedings of the International Conference on Machine Learning (ICML)}, 2021.

\bibitem{schaeffer2021algorithmic}
R.~Schaeffer, ``An algorithmic theory of metacognition in minds and machines,'' arXiv:2111.03745, 2021.

\bibitem{vaswani2017attention}
A.~Vaswani \emph{et al.}, ``Attention is all you need,'' \emph{Advances in Neural Information Processing Systems}, vol.~30, 2017.

\end{thebibliography}
\end{document}